\documentclass[a4paper]{article}
\usepackage{template}
\usepackage{multirow}
\usepackage{url}
\usepackage{footnote}

\title{Analyzing Effect of Repeated Reading on Oral Fluency and Narrative Production for Computer-Assisted Language Learning}
\name{Santosh Kumar Barnwal\textsuperscript{*}$^1$, Uma Shanker Tiwary$^1$}

\address{\textsuperscript{*}ORCiD: 0000-0002-2679-7797\\
  $^1$SILP Lab, Indian Institute of Information Technology, Allahabad}
\email{iis2009002@gmail.com, ustiwary@gmail.com}

\begin{document}

\maketitle
\begin{abstract}
Repeated reading (RR) helps learners, who have little to no experience with reading fluently to gain confidence, speed and process words automatically. The benefits of repeated readings include helping all learners with fact recall, aiding identification of learners' main ideas and vocabulary, increasing comprehension, leading to faster reading as well as increasing word recognition accuracy, and assisting struggling learners as they transition from word-by-word reading to more meaningful phrasing. Thus, RR ultimately helps in improvements of learners' oral fluency and narrative production. However, there is no open audio datasets available on oral responses of learners based on their RR practices. Therefore, in this paper, we present our dataset, discuss its properties, and propose a method to assess oral fluency and narrative production for learners of English using acoustic, prosodic, lexical and syntactical characteristics. The results show that a CALL system can be developed for assessing the improvements in learners' oral fluency and narrative production.
\end{abstract}

\section{Introduction}

Due to emergence as global language, English as a second language has been established as primary medium of instruction in higher education in several developing countries including India. Reading in a second language (L2) differs from reading in a first language (L1) in distinct ways. L1 learners have well-developed oral proficiency, vocabulary knowledge, and tacit grammar knowledge at the time they start learning to read, which leads to fluent processing of text information. L2 learners have limited oral proficiency and learners, and underdeveloped grammar knowledge. Therefore, compared to L1 learners, L2 learners are invariably slower and less accurate in processing text. One highly recommended procedure for improving L2 learners' oral fluency is learners readings \cite{gorsuch2015repeated}. 

Researchers of different domains such as prosody, acoustics, lexicon, syntactics proposed several measures to evaluate L2 learners' oral proficiency and narrative production. These measures can be applied for evaluating the impact of learners' repeated readings on their proficiency improvement. Also, availability of computational tools belonging to above mentioned domains make possible to develop a CALL system for L2 learners. A Computer Assisted Language Learning (CALL) system enables convenient and low-cost language learning, which focuses on developing the speaking, listening, and writing skills, and some of them are put to practice \cite{Wu2018}.

\subsection{Repeated Reading}
RR requires learners to reread a passage several times to achieve a pre-established level of fluency. The goals for RR are to increase learners' reading speed, transfer learning to new passages, and improve comprehension. There are two forms of repeated reading: unassisted and assisted. With unassisted repeated reading, learners are given reading passages that contain recognizable words at their independent reading levels. Each learner silently or orally reads his/her passage several times until he/she reaches the predetermined level of fluency. Assisted repeated reading, on the other hand, involves repeated reading whilst or after listening to either a teacher reading the same text or a recorded version \cite{liu2016implementation}. RR has been found to increase fluency and comprehension for first-language (L1) \& second-language (L2) learners not only with treatment texts but with new, unpracticed learners \cite{gorsuch2015repeated}. Thus, RR leads to improvements in speech prosody, a component of reading fluency indicative of learners' comprehension of texts \cite{ardoin2016repeated}.

\subsection{Acoustics-Prosody}
Prosody describes variation in intonation, duration, rhythm, and intensity, is a critical component of perceived fluency in spoken language, as prosodic variation signals not only syntactic and semantic structure of sentences but also emotion. For example, Kuhn et al. (2010) stated `in addition to the role of rate and accuracy, prosodic fluency requires appropriate expression or intonation coupled with phrasing that allows for the maintenance of meaning' \cite{kuhn2010aligning}. 

Several researchers have assessed the relationship between prosody and acoustics, about how prosodically fluent learners cue syntactic structure and semantic structure. For example, speakers often cue syntactic phrase boundaries through the employment of intonational phrase boundaries, the presence of silence between words and a pitch excursion, which can be rising in interrogatives sentences or falling in declaratives. Imoto et al. (2002) addressed sentence-level stress detection of English for Computer-Assisted Language Learning by Japanese learners \cite{imoto2002modeling}. 
Trofimovich et al. (2006) used acoustic features (syllables count, variation in stress, pitch rise and fall, and duration) to determine how accurately five prosody features (stress timing, peak alignment, speech rate, pause frequency, and pause duration) were produced by L2 English speakers \cite{trofimovich2006learning}. 

\subsection{Lexical richness}
Horst et al. (2000) found that repeated readings of L2 text help learners to identify the meaning and form of words without access to a dictionary or other learning support \cite{horst1999test}. With phonological support, it provides a supportive environment for both incidental and intentional novel vocabulary acquisition. Many L2 development studies have reported that a variety of lexical richness measures, along with measures of accuracy, fluency, and grammatical complexity, can be used as reliable and valid indices of the learner's developmental level or overall proficiency in an L2 \cite{lu2012relationship}. 

\subsection{Syntactic complexity}
Ortega (2003) stated `Syntactic complexity (syntactic maturity or linguistic complexity) refers to the range of forms that surface in language production and the degree of sophistication of such forms' \cite{ortega2003syntactic}. The measures used to examine syntactic complexity in L2 English writing development include length of production unit (e.g., T-unit, clauses, verb phrases, and sentences), amount of embedding, subordination and coordination, range of structural types, and structural sophistication. Several studies have examined relations between the syntactic complexity of speech and the speakers' holistic speaking proficiency levels. Iwashita's (2006) study on Japanese L2 speakers found that length-based complexity features (i.e., number of T-units and number of clauses per T-unit) are good predictors for oral proficiency \cite{iwashita2006syntactic}. To realize a voice-interactive CALL system, Anzai et al. proposed n-gram model based methods for improving recognition accuracy of speech with grammatical mistakes \cite{anzai2012recognition}.

\section{Collection of Repeated Reading data}
\subsection{Experimental procedure}

The purpose of this study was to develop a CALL system to examine the effectiveness of an repeated reading practice that is used without corrective feedback on the fluency and comprehension of narrative and expository articles for L2 English Indian students. The design of data collection experiment took inspiration from Sukhram et al. (2017) experiment  \cite{sukhram2017effects}. \\
\textbf{Participants:} 20 undergraduate students (7 females and 13 males, mean age = 20.29 years) pursuing engineering degree were participated for course credit. They all performed academic activities in English only, whereas their primary languages were different.\\
\textbf{Materials:} As all participants were studying with different subjects, therefore two articles were selected from the outside of their academic curriculum; article-1 was a simple narrative story `How the Camel got his Hump' by Rudyard Kipling (736 words) which were taken from the English textbook `It So Happened'\footnote{\url{http://ncertbooks.prashanthellina.com/class_8.English.ItSoHappend/index.html}} Class 8th NCERT, whereas article-2 was a relatively more complex expository article `Nineteenth-century politics in the United States'\footnote{\url{http://toeflpreparationsources.weebly.com/uploads/1/2/5/1/12516531/reaing_test_-_text_1.pdf}} (668 words) selected from reading section of TOEFL-iBT test guide book. Both articles were unread in their life span until the experiment. These articles were displayed on a monitor without any title, heading, subheading, and illustration. \\
\textbf{Procedures:} The experiment was held in a research lab for three consecutive days, where participants were tested in a quiet room.
On day 1, each participant sat in front of a monitor that was displaying the article-1 to read. Mostly participants preferred silent reading, therefore, their oral reading could not be recorded. There was no time limit for reading to provide natural reading conditions. 
Therefore, participants read at their own pace. After reading the article, students were instructed to speak summary in as great a detail as possible, without referring back to the article. Their speech were recorded using a digital audio recorder software. Students then moved on to reading the article-2 using the same procedures described above. Same experiments were conducted on day 2, and 3, where same participants were reading same both articles and after their reading, summary-speech were recorded.

The recordings were transcribed in English at word-level by manual correction of a transcript generated automatically using the Google Speech API recognizer \cite{google}, where brief pauses were marked with commas, while long pauses were marked with full stops (end of sentence) if their places were according to semantic, syntactic and prosodic features. All disfluencies in the speech (mispronunciations, hesitations, repetitions, repairs, deletion, substitution, insertion, incomplete and incomprehensible sounds) were maintained separately. Also, their corresponding words were not included in final transcripts.

\subsection{Evaluation of collected data}

For the annotation of the data, we employed three independent raters who have experience of teaching English to Indian students. The evaluators annotated all summary speech with the following four criteria. \\
\textbf{Oral fluency: } How natural the student's pronunciation, rhythm and intonation of the speech.\\
\textbf{Lexical richness: }How students applied lexicons of articles in their sentence production.\\
\textbf{Syntactic maturity: }How effective the student's syntactic maturity in their narrative production.\\
\textbf{Overall score: }How proficient the student's language production is. 

Each criterion was evaluated by three-point scale and given labels: basic, average, advance. Inter-rater agreement was 93\% for fluency, 90\% for lexical richness, 85\% for syntactic maturity and 89\% for overall score.

\begin{table*}[th]
\tabcolsep 2pt
\small
  \caption{Accuracy (\%) of various classifiers on different prosodic feature-sets}
  \label{tab:example}
  \centering
  \begin{tabular}{lcccccc}
    \toprule
    \textbf{Feature Set} & \multirow{2}{*}{\textbf{Day}} & \multirow{2}{*}{\textbf{SVM}} & \textbf{Logistic} & \textbf{Nearest} & \textbf{Decision} & \textbf{Random}\\
    \textbf{(\#Features)} &  &  & \textbf{Regression} & \textbf{Neighbors} & \textbf{Trees} & \textbf{Forest} \\
    \midrule
    \multirow{3}{*}{\textbf{Avec2013 (2268)}} & 1 & 84.1 & 86.15 & 73.85 & 75.9 &83.08 \\
    &2&82.76&85.82&83.14&81.23&83.91\\
	&3&84.23&80.87&82.55&74.83&83.89\\
	\midrule
    
        \multirow{3}{*}{\textbf{ComParE2016 (6373)
}}&1&80.51&83.59&69.74&70.26&78.97\\
&2&81.61&\textbf{86.97}&77.01&81.23&80.84\\
&3&77.52&81.21&77.85&74.5&79.87\\
	\midrule
	    
	    \multirow{3}{*}{\textbf{eGeMAPS (88)
}}&1&72.82&73.33&72.82&67.18&76.92\\
&2&83.91&84.29&84.67&78.16&84.67\\
&3&82.21&80.2&83.22&80.54&83.56\\
	\midrule
	
	    \multirow{3}{*}{\textbf{IS09 emotion (384)
}}&1&81.54&75.9&70.26&61.03&74.87\\
&2&81.99&79.69&78.93&80.08&79.31\\
&3&80.6&81.61&75.59&72.91&78.6\\
	\midrule
	
		\multirow{3}{*}{\textbf{IS10 paraling (1582)
}}&1&\textbf{87.18}&85.64&83.59&70.77&81.54\\
&2&83.91&82.38&84.67&79.69&83.91\\
&3&\textbf{84.56}&80.87&83.89&75.5&82.21\\
	\midrule
	
		\multirow{3}{*}{\textbf{IS11 speaker state (4368)
}}&1&77.95&85.64&69.23&72.82&78.97\\
&2&80.08&86.59&78.54&75.48&80.46\\
&3&78.86&79.19&78.52&71.48&78.19\\
	\midrule
	
		\multirow{3}{*}{\textbf{IS12 speaker trait (5757)
}}&1&80.51&84.1&68.21&74.87&77.95\\
&2&82.38&86.21&77.78&81.99&82.38\\
&3&78.86&80.2&79.19&72.82&79.53\\
	\midrule
	
		\multirow{3}{*}{\textbf{IS13 ComParE (6373)
}}&1&76.92&84.1&64.62&71.28&81.03\\
&2&81.61&86.97&77.01&84.29&80.84\\
&3&77.85&80.2&77.18&72.48&79.87\\		
    \bottomrule
  \end{tabular}
  
\end{table*}

\section{Features Extraction and Analysis}
We describe the implementation of acoustic-prosodic features from speech, as well as lexical and syntactic features from corresponding transcript. 
\subsection{Acoustic-Prosodic Features}
In order to increase the number of speech segments for acoustic-prosodic analyses, we segmented the speech audio obtained above into 0.5 seconds chunks (termed fragments). For each summary speech, respective silence to speech ratios was obtained from the total duration of silence and speech. After that, all fragments containing silence only were removed from the analysis process.

Acoustic low-level descriptors (LLD) and temporal features were extracted from speech using the openSmile toolkit (Eyben et al., 2013) \cite{eyben2010opensmile}. We used the feature sets specified in the extended Geneva Minimalistic Acoustic Parameter Set (eGeMAPS)\cite{eyben2016geneva}, the Continuous Audio-Visual Emotion and Depression Recognition Challenge (avec2013)\cite{valstar2013avec}, INTERSPEECH 2016 Computational Paralinguistics Challenge (ComParE2016)\cite{schuller2016interspeech}, Interspeech 2009 Emotion Challenge (IS09 emotion)\cite{schuller2009interspeech}, INTERSPEECH 2010 Computational Paralinguistics Challenge (IS10 paraling)\cite{schuller2013paralinguistics}, Interspeech 2011 Speaker State Challenge (IS11 speaker state)\cite{schuller2011interspeech}, INTERSPEECH 2012 Speaker Trait Challenge (IS12 speaker trait)\cite{schuller2012interspeech}, and INTERSPEECH 2013 Computational Paralinguistics Challenge (IS13 ComParE)\cite{schuller2013interspeech}. Previous prosodic and paralinguistic researchers used these feature sets in various related studies, such as a) speech emotion recognition \cite{manfeature} , b) assessment of sincerity in speech \cite{albornoz2016design}, c) native language identification \cite{shivakumar2016multimodal}, d) speech impairment analysis \cite{corrales2018acoustic}, e) language proficiency assessment \cite{zhang2016language}, f) personality recognition \cite{an2018lexical}, g) affect identification \cite{park2018using}. 

The LLD features include pitch (fundamental frequency), intensity (energy), spectral, cepstral (MFCC), duration, voice quality (the zero-crossing rate, jitter, shimmer, and harmonics-to-noise ratio), spectral harmonicity, and psychoacoustic spectral sharpness. 
	
\subsection{Lexical richness}
In the language acquisition literature, measures of lexical richness were grouped in three interrelated components: lexical density, sophistication, and variation \cite{laufer1995vocabulary}. \\
\textbf{Lexical density: }It refers to the ratio of the number of lexical words to the total number of words in a text. lexical words referred as nouns, adjectives, verbs (excluding modal verbs and auxiliary verbs), and adverbs with an adjectival base (e.g., “particularly”).\\
\textbf{Lexical sophistication: }It refers to the ratio of relatively unusual or advanced words in the learner's text.  \\
\textbf{Lexical variation: }It refers to the range of a learner's vocabulary as displayed in his or her language use. Some of important indices are: number of different words in a text, Type–token ratio (TTR), i.e., the ratio of the number of word types (T) to the number of words (N) in a text, mean segmental TTR, corrected TTR, and root TTR.

Lu (2012) examined the relationship of lexical richness to the quality of L2 English learners' oral narratives. He designed a computational system to automate the measurement of above mentioned components of lexical richness using 25 different metrics \cite{lu2012relationship}. In the present work, we used these metrics to measure the improvements in L2 student's fluency and narrative production.

\begin{table}[th!]
\tabcolsep 2pt
  \caption{Accuracy (\%) of various classifiers on Lexical Richness}
  \label{tab:example}
  \centering
  \begin{tabular}{lccccc}
    \toprule
    \multirow{2}{*}{\textbf{Cases}} & \multirow{2}{*}{\textbf{SVM}} & \textbf{Logistic} & \textbf{Nearest} & \textbf{Decision} & \textbf{Random}\\
    &  & \textbf{Regression} & \textbf{Neighbors} & \textbf{Trees} & \textbf{Forest} \\
    \midrule
\textbf{All}&83.87&77.42&\textbf{90.32}&74.19&87.1\\
\textbf{Day-1}&63.64&\textbf{72.73}&54.5&63.64&63.6\\
\textbf{Day-2}&72.73&72.7&63.64&\textbf{90.91}&81.82\\
\textbf{Day-3}&54.55&36.36&54.4&\textbf{81.82}&63.64\\
\textbf{Article-1}&81.25&75&75&93.7&\textbf{100}\\
\textbf{Article-2}&81.61&75&62.5&81.2&\textbf{100}\\
    \bottomrule
  \end{tabular}
  
\end{table}

\begin{table}[th!]
\tabcolsep 2pt
  \caption{Accuracy (\%) of various classifiers on Syntactic Features}
  \label{tab:example}
  \centering
  \begin{tabular}{lccccc}
    \toprule
    \multirow{2}{*}{\textbf{Cases}} & \multirow{2}{*}{\textbf{SVM}} & \textbf{Logistic} & \textbf{Nearest} & \textbf{Decision} & \textbf{Random}\\
    &  & \textbf{Regression} & \textbf{Neighbors} & \textbf{Trees} & \textbf{Forest} \\
    \midrule
\textbf{All}&74.19&67.74&64.52&74.19&77.42\\
\textbf{Day-1}&72.7&63.64&63.6&81.8&72.7\\
\textbf{Day-2}&45.5&54.55&45.4&45.4&45.45\\
\textbf{Day-3}&63.64&81.2&72.7&72.7&63.64\\
\textbf{Article-1}&81.25&68.7&68.75&81.25&87.5\\
\textbf{Article-2}&75&56.25&75&87.5&81.25\\
    \bottomrule
  \end{tabular}
  
\end{table}

\begin{table}[th!]
\tabcolsep 2pt
  \caption{Accuracy (\%) of various classifiers on fused Features of Prosody, Lexical and Syntactic properties}
  \label{tab:example}
  \centering
  \begin{tabular}{lccccc}
    \toprule
    \multirow{2}{*}{\textbf{Cases}} & \multirow{2}{*}{\textbf{SVM}} & \textbf{Logistic} & \textbf{Nearest} & \textbf{Decision} & \textbf{Random}\\
    &  & \textbf{Regression} & \textbf{Neighbors} & \textbf{Trees} & \textbf{Forest} \\
    \midrule
\textbf{All}&70.8&61.29&64.52&58.06&\textbf{74.19}\\
\textbf{Day-1}&63.64&54.5&45.45&54.5&\textbf{72.7}\\
\textbf{Day-2}&\textbf{81.82}&\textbf{81.82}&\textbf{81.82}&54.5&\textbf{81.8}\\
\textbf{Day-3}&\textbf{81.82}&63.64&\textbf{81.82}&72.7&63.6\\
\textbf{Article-1}&87.5&75&75&62.5&81.25\\
\textbf{Article-2}&\textbf{68.75}&56.25&56.25&\textbf{68.75}&\textbf{68.75}\\
    \bottomrule
  \end{tabular}
  
\end{table}

\subsection{Syntactic analysis}
Pallotti (2015) stated `syntactic complexity can be measured by looking at the number of interconnected constituents in a structure, which is the principle behind three measures such as length of phrase, number of phrases per clause and number of clauses per unit' \cite{pallotti2015simple}. Lu (2010) described a computational system for automatic analysis of syntactic complexity in L2 English writing to measure following five components using fourteen different metrics: \textbf{Length of production unit}- mean length of clause (MLC), mean length of sentence (MLS), and mean length of T-unit (MLT); \textbf{Sentence complexity}- clauses per sentence (C/S); \textbf{Subordination}- clauses per T-unit (C/T), complex T-units per T-unit (CT/T), dependent clauses per clause (DC/C), and dependent clauses per T-unit (DC/T); \textbf{Coordination}- coordinate phrases per clause (CP/C), coordinate phrases per T-unit (CP/T), and T-units per sentence (T/S); and \textbf{Particular structures}- complex nominals per clause (CN/C), complex nominals per T-unit (CN/T), and verb phrases per T-unit (VP/T) \cite{lu2010automatic}. We used these metrics for complexity computation on speech transcripts as suggested in \cite{chen2011computing}.

\begin{figure}[t]
  \centering
  \includegraphics[width=\linewidth]{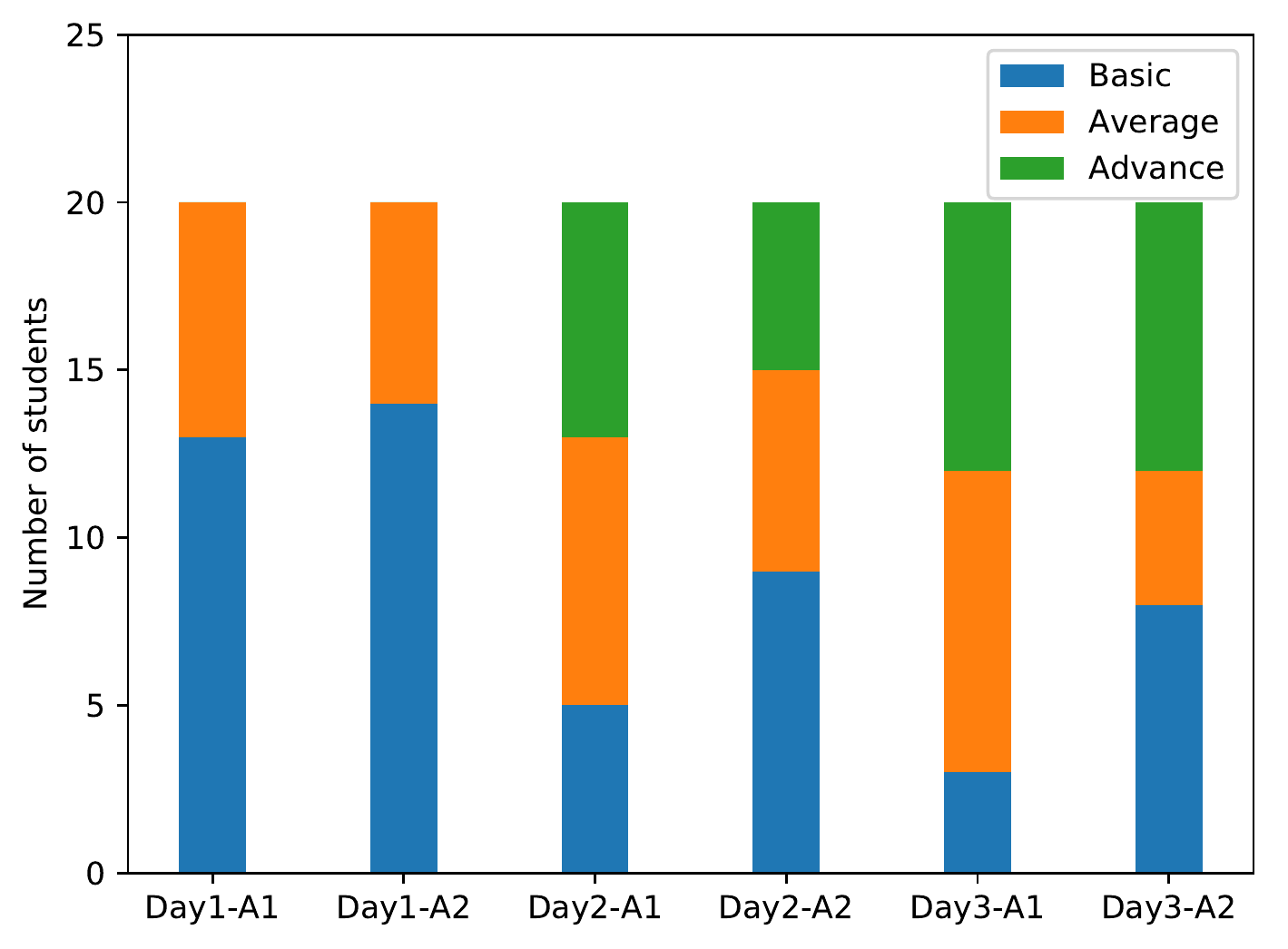}
  \caption{Stack bar of the average of ratings given by three raters for overall criterion.}
  \label{fig:rating}
\end{figure}

\section{Results and Discussion}
In this section, we measure performance of the proposed system on four criteria discussed in section 2. For each criterion, features of dataset were splited into training (70\%) and testing (30\%) as well as labelled in three categories: basic, average, and advance. These labels were calculated by averaging the raters' scores. Five popular supervised learning classifiers: SVM, Logistic regression, Nearest Neighbors, Decision trees and Random forests were applied on the dataset.\\
\textbf{Oral fluency evaluation: }Table 1 presents accuracy results of the oral fluency classification on five classifiers with various feature-sets.  The results show that for days 1 and 3, SVM classifier provided highest accuracy: 87.18\% and 84.56\% with IS10 paraling, whereas, for day 2, logistic regression classifier's accuracy was the highest with ComParE2016. Minimum accuracies of all classifiers were above 61.0\% for all three days. \\
\textbf{Lexical richness: } The performance of classifiers on lexicon features were reported in Table 2. First row shows the Nearest neighbors classifier provided highest accuracy on whole lexicon features. Next three rows show the Decision trees classifier provided best results for days 2 and 3, whereas Logistic regression is best for day 1 lexicon features. Last two rows show that Random Forest outperformed on article based lexicon features. \\
\textbf{Syntactic analysis: }The performance of classifiers on syntactic features were reported in table 3. First row shows the Random forest classifier provided highest accuracy on whole syntactic features. Next three rows show the Logistic Regression provided best results for days 2 and 3, whereas Decision trees was best for day 1 syntactic features. Last two rows show the Random Forest and Decision trees provided the best accuracy on syntactic features of article 1 and 2 respectively.\\
\textbf{Overall Analysis: }Overall features were estimated by early fusion of lexicon, syntactic and prosody features. Lexicon and syntactic features were the same as discussed earlier. Prosody features were extracted by applying eGeMAPS on speech without splitting into fragments. This process helps in preventing the domination of prosody features over lexicon and syntactic features in training. The performance of classifiers on fused features were reported in Table 4. The Random forest and SVM classifiers delivered the best accuracy in comparison to others. Repeated reading effect on overall score is shown in figure~\ref{fig:rating}, number of students labelled as 'advance' raised on both articles (A1 and A2) as increasing reading practices on days 1, 2, and 3.

\section{Conclusion}
In this research, we constructed the dataset and analysed the effect of repeated readings on learners' performance based on four criteria: oral fluency, lexical richness, syntactic maturity and overall score. To measure oral fluency, we extracted acoustic-prosodic features by applying various feature-sets. Lexicon and syntactic features were extracted from transcripts. We proposed methods to classify the learners' performance into three catogories: basic, average, and advance. We analysed the methods in three variations : days, articles, and both. We found that the proposed methods can track the improvements in the learners' oral fluency and narrative production. The methods also provide the facilities to compare individual's learning rate with others. 
The accuracies of five classifiers demonstrate the feasibility of developing an automatic system to evaluate learners' performance. 
In future work, we plan to expand the dataset by increasing the number of the participants as well as adding diverse reading exercises on various text. The analysis of disfluencies in speech considered to be studied further in order to improve performance of the proposed system.

\bibliographystyle{IEEEtran}

\bibliography{template}

\end{document}